\documentclass[10pt,twocolumn,letterpaper]{article}

\usepackage{cvpr}      
\usepackage{tabularx}
\usepackage{graphicx}
\usepackage{multirow}
\usepackage{etoolbox}  

\usepackage{Commands}

\newcommand{\class}{C}
\newcommand{\name}{n}

\definecolor{cvprblue}{rgb}{0.21,0.49,0.74}
\usepackage[pagebackref,breaklinks,colorlinks,allcolors=cvprblue]{hyperref}

\def\paperID{41} 
\def\confName{EarthVision}
\def\confYear{2025}

\title{Visual Question Answering on Multiple Remote Sensing Image Modalities}
\author{
    Hichem Boussaid\textsuperscript{1,*,†}, 
    Lucrezia Tosato\textsuperscript{1,2,*,†}, 
    Flora Weissgerber\textsuperscript{2} \\
    Camille Kurtz\textsuperscript{1}, 
    Laurent Wendling\textsuperscript{1}, 
    Sylvain Lobry\textsuperscript{1,†}\\
    \small
    \textsuperscript{1}LIPADE, Université Paris Cité, France \\
    \small
    \textsuperscript{2}ONERA, France\\
    \small
    \textsuperscript{†}corresponding authors: \{hichem.boussaid, lucrezia.tosato\}@etu.u-paris.fr
}

\begin{document}
\maketitle
\begin{abstract}
The extraction of visual features is an essential step in Visual Question Answering (VQA). Building a good visual representation of the analyzed scene is indeed one of the essential keys for the system to be able to correctly understand the latter in order to answer complex questions.
In many fields such as remote sensing, the visual feature extraction step could benefit significantly from leveraging different image modalities carrying complementary spectral, spatial and contextual information. 
In this work, we propose to add multiple image modalities to VQA in the particular context of remote sensing, leading to a novel task for the computer vision community. 
To this end, we introduce a new VQA dataset, named TAMMI (\textit{Text and Multi-Modal Imagery}) with diverse questions on scenes described by three different modalities (very high resolution RGB, multi-spectral imaging data and synthetic aperture radar). 
Thanks to an automated pipeline, this dataset can be easily extended according to experimental needs.
We also propose the MM-RSVQA (\textit{Multi-modal Multi-resolution Remote Sensing Visual Question Answering}) model, based on VisualBERT, a vision-language transformer, to effectively combine the multiple image modalities and text through a trainable fusion process. 
A preliminary experimental study shows promising results of our methodology on this challenging dataset, with an accuracy of \textbf{65.56\%} on the targeted VQA task.
This pioneering work paves the way for the community to a new multi-modal multi-resolution VQA task that can be applied in other imaging domains (such as medical imaging) where multi-modality can enrich the visual representation of a scene. The dataset and code are available at \href{https://tammi.sylvainlobry.com/}{https://tammi.sylvainlobry.com/}.
\begingroup
\renewcommand\thefootnote{}\renewcommand\footnotemark{}\footnotetext{\hspace{-5mm}*Hichem Boussaid and Lucrezia Tosato contributed equally.}
\renewcommand\thefootnote{}\renewcommand\footnotemark{}\footnotetext{\hspace{-5mm}This work is supported by \textit{Agence Nationale de la Recherche} (ANR) under the ANR-21-CE23-0011 project. The experiments conducted in this study were performed using HPC/AI resources provided by GENCI-IDRIS (Grant 2023-AD011012735R2).}
\endgroup
\end{abstract}  

\begin{figure}
\centering
\includegraphics[width=\columnwidth]{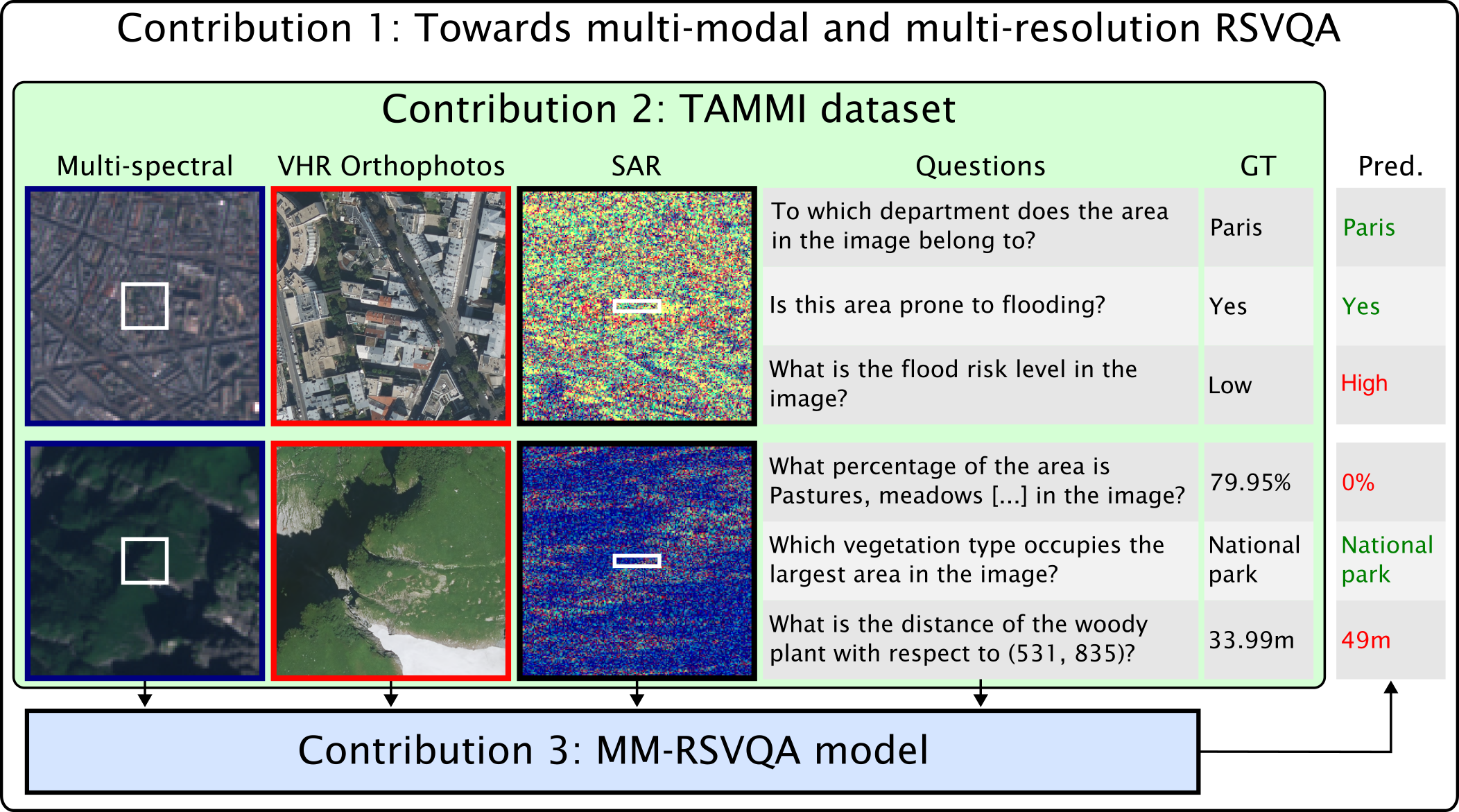}
\caption{Summary of our contributions.
We introduce a new task in the computer vision community with multi-modal and multi-resolution Visual Question Answering (VQA) on remote sensing images. 
We introduce a new dataset, TAMMI, associating question/answer pairs to multi-spectral, Very High-Resolution (VHR) orthophotos and Synthetic Aperture Radar (SAR) images triplets. 
In these examples, the white rectangle in the multi-spectral and SAR images corresponds to the extent of the VHR image. 
Finally, we propose a new model for this task referred as MM-RSVQA.}
\label{fig:intro_QA}
\end{figure}

\section{Introduction}
\label{sec:intro}
The task of Visual Question Answering (VQA) aims at providing natural language answers to free-form, open-ended question about an image~\cite{antol2015vqa}. 
On natural images, recent advances in the computer vision and natural language processing communities have shown great improvements to standard VQA benchmarks. 
In particular, Large Language Models (LLM) can now be used to perform knowledge-based VQA, which requires external knowledge and commonsense reasoning~\cite{hu2023promptcap}. 
These models are however more limited when used beyond natural images~\cite{li2023comprehensive, khan2023q}.

VQA has been proposed for the medical domain~\cite{hasan2018overview} and is an active field of research in the medical and computer vision communities~\cite{lin2023medical}. 
The VQA task is also used for extracting information from remote sensing data~\cite{lobry2020rsvqa}. 
In such thematic domains, there are significant challenges compared to VQA for natural images. 
In particular, the limited availability of high-quality training data and the variability of information found in different imaging modalities have been active research topics~\cite{lin2023medical, yuan2023multilingual, khan2023q}.

It is commonly agreed that VQA methods for both of the remote sensing and medical imaging fields would benefit from information contained in different modalities of images~\cite{persello2022deep}. 
Indeed, multiple image modalities are often used in these fields to obtain different, specific and complementary (spectral, spatial and contextual) information about a single scene.
In remote sensing in particular, multi-modal data allows to obtain different information at multiple resolutions and from various wavelengths, from the 400nm of ultraviolet to the 5cm of radar wave. 
Among others, Very-High Resolution (VHR) data offers images at sub-meter resolutions. 
With Multi-Spectral (MS) data, it is possible to characterize different ground materials thanks to their different spectral responses. 
Synthetic Aperture Radar (SAR) offers imaging capacities highlighting man-made objects and giving physical information about objects~\cite{li2022deep}.
While previous works have explored VQA from multi-modal images \cite{tosato2024can}, using RGB data from the Sentinel-2 satellites and SAR data from Sentinel-1 on existing datasets, dedicated datasets and methods are necessary. 

Our first contribution (highlighted in \autoref{fig:intro_QA}) is to tackle the task of VQA from multi-modal and multi-resolution images leading to a novel challenging task for the computer vision community. 
Our second contribution is a new dataset named TAMMI, built from openly available data sources. 
This dataset combines three modalities (VHR, MS and SAR) and we make it openly available. 
We also share with the vision community an automated pipeline to easily extend this dataset (e.g. to new geographical areas or modalities) according to experimental needs.
MS and SAR can be used dynamically at different sizes to provide different levels of context. 
Finally, our third contribution is a baseline model for the targeted multi-modal VQA task named MM-RSVQA. 
The proposed architecture is able to take as input the three image modalities, with specifically pre-trained feature extractors. 
To effectively integrate the multi-modal visual features, along with the textual features of the question, we use VisualBERT, a recent vision-language transformer, leading to a trainable fusion scheme. 

\section{Related works}
\label{sec:SoA}
\textbf{VQA} has the objective of predicting a natural language response to an open-ended question related to an image~\cite{antol2015vqa}. 
This task bridges visual reasoning with semantics expressed in natural language, providing new challenges for the computer vision community. 
Models relying on attention~\cite{yang2016stacked, anderson2018bottom, zhou2021trar} to extract relevant features have been proposed. 
Recently, foundation models such as CLIP~\cite{radford2021learning} have shown remarkable success on the VQA downstream task~\cite{shen2021much}.
However, such approaches do not translate directly to thematic applications of the VQA task~\cite{eslami2023pubmedclip}.

The task of VQA for remote sensing (RSVQA) is introduced in~\cite{lobry2020rsvqa}. 
In this work, features are extracted using a Convolution Neural Network (CNN) and a Recursive Neural Network (RNN) and fused together with a point-wise multiplication. 
Improvements in the feature extractor have been proposed by Felix et al.~\cite{felix2021cross}, inspired by LXMERT~\cite{tan2019lxmert} enhancing the results using an object detection step implemented via the Faster-RCNN model as a visual encoder and BERT for the language part. 
The Fourier transform is also used in~\cite{zhaofrequency} to extract structural information from complex remote sensing data, thereby improving the generalizability across various domains.
In~\cite{chappuis2023multi}, an object detector and a classifier are used to create a caption of the image, then used to answer the question. 
LLMs such as BERT~\cite{devlin2018bert} have been used for textual features extraction~\cite{chappuis2022language} or as the main component of the model~\cite{chappuis2022prompt}. 
The fusion step of the two representations is done in~\cite{felix2021cross} with a cross-modal transformer encoder. 
Another attention-based method is proposed in~\cite{zheng2021mutual}, with a fusion module based on mutual attention. 
In~\cite{tosato2024segmentation}, a method that uses segmentation maps to guide the attention is introduced. 
In parallel, training techniques such as self-supervised curriculum learning~\cite{yuan2021self} have been shown to be efficient to obtain a common representation of textual and visual features. 
Improvements to the language processing have been introduced as well: in~\cite{yuan2023multilingual} an augmentation is applied to translate each question in multiple languages and then translate it back to English, improving the diversity of the formulation of questions. 

\noindent\textbf{Datasets} Several datasets have been created for RSVQA. 
This problem is first considered in~\cite{lobry2020rsvqa} which introduced two RSVQA datasets composed of image/question/answer triplets. 
The first one, called "Low Resolution (LR)", is based on images from the Sentinel-2 sensors (optical images with a spatial resolution of 10m) acquired over The Netherlands. 
The second one, called "High Resolution (HR)", uses RGB aerial images with a spatial resolution of 15cm extracted from the USGS High Resolution Orthoimagery database, covering urban areas of the United States. 
In these two datasets, the nature of the questions, as well as the distribution of answers, is highly unbalanced (e.g. ’0’ in the HR dataset has a frequency of 60.9\% for the numerical answer). 
In addition, the number of different answers is very limited, with 9 possible answers for LR and 98 for HR. 
RSVQAxBEN~\cite{lobry2021rsvqa} dataset proposes a larger number of samples and introduces new objects of interest (land cover classes) with a new form of complexity (logical formulas). 
The area of interest is also different, as it covers many European countries thanks to images from the Sentinel-2 satellites.  
However, even in this dataset, the imbalance in the distribution of answers remains. 
In order to increase the diversity of questions, Zheng et al.~\cite{zheng2021mutual} have exploited five pre-annotated datasets, three for scene classification and two for object detection for the automatic construction of questions and answers.

These datasets have certain limitations in common. 
First, most of them have a limited number of samples, which may reduce the ability of deep learning models to generalise effectively~\cite{lobry2020rsvqa}. 
In addition, the diversity of questions and answers is often restricted, which can lead to biases and gaps in model performances~\cite{chappuis2023curse}. 
Finally, these datasets focus on the use of RGB images, not leveraging other modalities, sensors and resolutions.

Integrating SAR imagery with natural language remains under-explored and challenging due to its unique data structure. Deep learning models often struggle with SAR-based tasks requiring precise quantification, such as target size estimation or object counting~\cite{zhao2022exploring}. However, they are effective in tasks involving spatial relationships, like proximity identification or density assessment within SAR scenes. SAR has been used in RSVQA for tasks such as scattering pattern classification~\cite{aghababaei2024visual}, ship detection and counting~\cite{wang2024visual}, and in combination with optical images for land cover related questions~\cite{tosato2024can}.


In this work, beyond sharing a new vision task with the community, we address existing gaps in the current state of the art by creating a diverse dataset that spans multiple complementary image modalities, resolutions, and geographic regions, encompassing a wide range of terrain types and incorporating various question typologies. 
As a first baseline on this task and dataset, we embed state of the art transformer-based fusion techniques proven effective through a new model for tackling multi-modal challenges.


\section{Dataset}
\label{sec:dataset}
\begin{figure}[!t]
\centering
\includegraphics[width=\columnwidth]{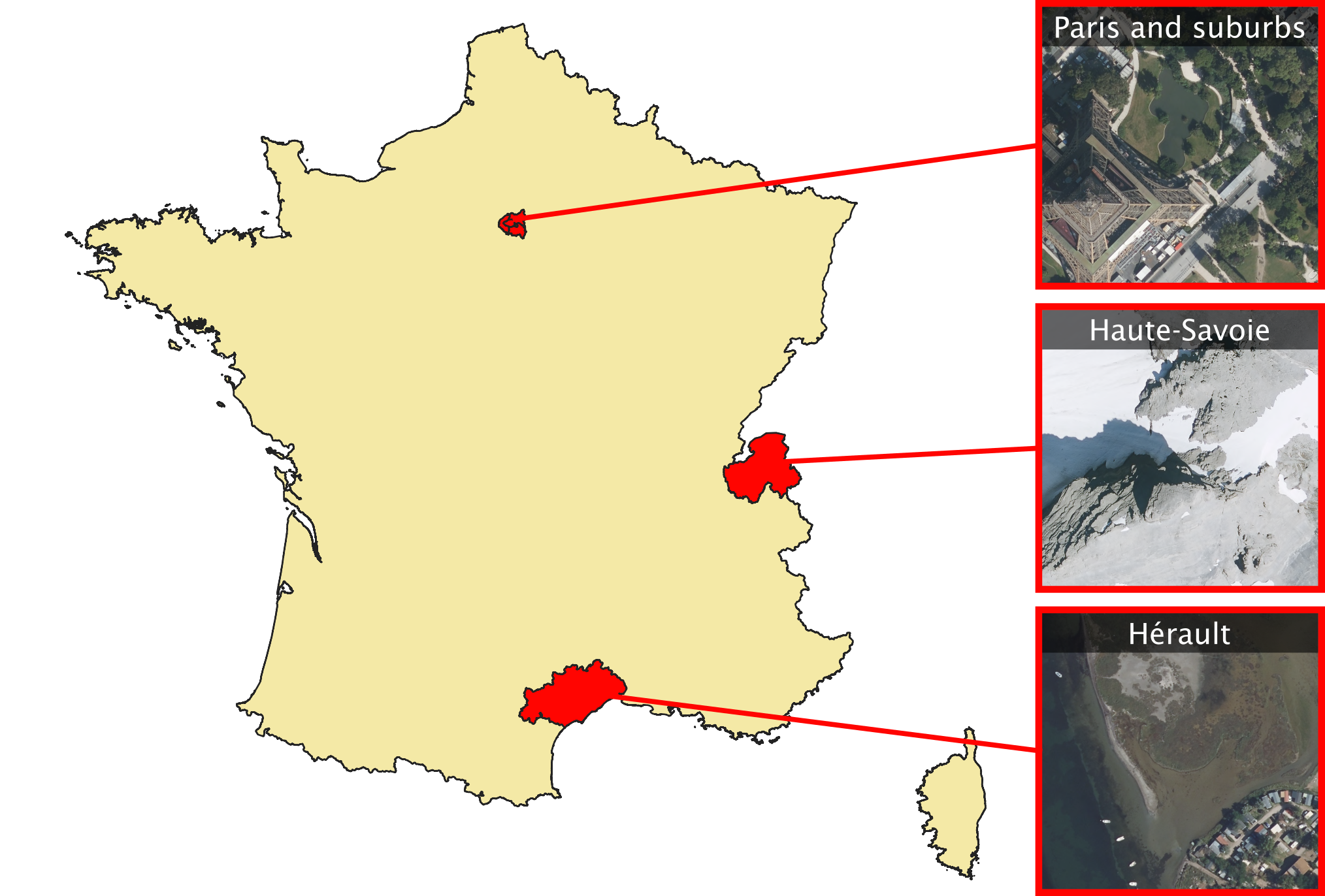}
\caption{Geographical extent of the TAMMI dataset, covering selected regions (highlighted in \textcolor{red}{red}) in Metropolitan France: 
Paris and inner suburbs (departments 75, 92, 93, 94), an urban region; 
Haute-Savoie (74), a mountainous region; 
and Hérault (34), a seaside region. 
For each of these regions, we show a VHR sample from the dataset.}
\label{fig:GeoExtent}
\end{figure}

For the purpose of advancing RSVQA capabilities we develop TAMMI (\textit{Text and Multi-Modal Imagery}), a large multi-modal and multi-resolution dataset. 
TAMMI includes images/question/answer triplets on three French departments, shown in \autoref{fig:GeoExtent}. 
The following subsections outline each phase of the dataset creation, detailing data sources and processing steps involved.

\subsection{Image modalities}
\paragraph{Very high-resolution orthophotos (BDOrtho)}
BDOrtho is a dataset of aerial VHR optical images acquired by the french National Geographic Institute (IGN). 
It provides an accurate and detailed photographic representation of the French territory at 20cm resolution. Images are provided in RGB and updated every three years. For our dataset, the most recent images are chosen for each department: images from department 74 are from 2020, and images for department 34, 75, 92, 93, 94 are from 2021. 
To ensure consistency in the information provided by the images, the same years are maintained for the images of the other modalities.

\paragraph{Multi-spectral data (Sentinel-2)}
Sentinel-2 is a mission of the European Union's Copernicus program, designed to provide multi-spectral images of land. 
Sentinel-2 captures data in 13 bands of the electromagnetic spectrum~\cite{S2mission}, including near-infrared (NIR), visible and short-wave infrared (SWIR) with a spatial resolution of 10m, 20m, or 60m depending on the band. 
Sentinel-2 images are acquired every five days when both Sentinel-2A and Sentinel-2B are active. 
The images are distributed as Level-1C (L1C) products which provide top-of-atmosphere (TOA) reflectance, and Level-2A (L2A) that offer surface reflectance after atmospheric correction. 
Since our dataset includes SAR images which are not affected by atmospheric conditions and BDOrtho images which are taken during sunny days and do not need any atmospheric correction, we select L1C products. 
Furthermore, only images with a cloud coverage under 3\% are selected. 

\paragraph{Synthetic Aperture Radar data (Sentinel-1)}
The Sentinel-1 satellite is also part of the European Union's Copernicus program,  designed to provide full coverage of Europe every 6 days.
Sentinel-1 acquires SAR images by sending radar pulses towards the earth's surface and measuring the backscattered signal. 
Since the radar pulses propagate through clouds, the acquisition rate is independent of the atmospheric conditions. 
In our dataset, we use Interferometric Wide (IW) swath mode. IW is the main acquisition mode over land and has a resolution of 5m in range (across satellite trajectory) and 20m in azimuth (along trajectory).
In this acquisition mode, the swath is divided in three sub-swathes. 
The satellite acquires tiles, called \textit{burst}, in each sub-swath sequentially with an overlap between the bursts. 
The bursts are processed as separate Single Look Complex (SLC) images. In the Sentinel-1 SLC products, sequentially acquired bursts of the same sub-swath are included into a single image separated by black bands.
Sentinel-1 acquires data in two polarization channels (VV and VH), providing additional information about the characteristics of materials on the planet surface.
In this work, we use the amplitude of the SLC level 1 images for both polarization channels.


\subsection{Image triplets}
To build the question/answer parirs, we select the geographical extent of the VHR patches $\pvhr$. The VHR patches are extracted from the division of the VHR tiles (of dimension 25000 $\times$ 25000 pixels) of our areas of interest (shown in \autoref{fig:GeoExtent}) into patches of size 1000 $\times$ 1000 pixels (i.e. 200m $\times$ 200m). The areas of interest are selected for their diverse landscapes. Specifically, the Île-de-France region (departments 75, 92, 93, 94) is predominantly urban, while the Haute-Savoie region (department 74) features mountainous terrain, and the Hérault region (department 34) is maritime. 
Since MS ($\pms$) and SAR ($\psar$) patches are designed to offer spatial context to the model, their extent (respectively $\lms$ and $\lsar$) is defined by the user and the patches are cropped at execution time. 
To align the images of all three modalities, we retrieve the latitude, longitude and altitude of the central pixel for each VHR patch. 
We then associate a MS and a SAR tile centered on the position of the central pixel of the VHR patch. 
The MS tiles are geocoded. Therefore, the latitude and longitude are converted to the image pixel space.
For SAR images, we use the method of~\cite{Weissgerber2022} to project the position of the central pixel of $\pvhr$.
To produce continuous SAR patches, the SAR images are debursted (removing black lines and burst overlaps), and pre-processed with tail-value removal based on histogram analysis for each polarization, then a transformation to decibels and a normalization are applied. The pre-processing steps are detailed in the supplementary materials.

\subsection{Question/answer pairs}
Inspired by \cite{johnson2017clevr}, we propose an automatic approach for the construction of question/answer pairs associated to each geographical area covered by the VHR patch, using: 
\begin{itemize}
    \item \textbf{BDTopo} provided by IGN (French Geographical Institute). It is an official vectorial description of the French territory, including generic geographical objects (e.g. buildings and water areas), and specific ones (e.g. museums and lakes);
    
    \item \textbf{Flooding risks database (TRI)} from \textit{Géorisques}. This dataset corresponds to the areas at significant risk of flooding, including management zoning, floodable area zoning, water level zoning, etc.;
    
    \item \textbf{Urban units 2020 (BU20)} of INSEE (French National Institute of Statistics and Economic Studies). This database corresponds to the urban zoning of French cities according to the continuity of the constructions and the number of inhabitants;
    
    \item \textbf{CORINE Land Cover (CLC) 2018} produced as part of the Copernicus European land monitoring program and managed by the European environment agency. It is a database of land cover and land use classifications, on three different levels of hierarchy with different levels of details. 
    In this work, we consider the Level-3 having the most detailed classes (44);
    
    \item \textbf{European Mountain Areas (EMA)} provided by the European environment agency. It contains information about the geometry, the area and the name of the mountains in Europe.
\end{itemize}

\paragraph{Construction of questions}
For a given VHR patch $\pvhr$, we retrieve the collection of geo-located objects $\objects$ that are present in $\epvhr$, the geographical extent of $\pvhr$.
A collection of objects $\mathbf{\objects_{\class}} = \{\objects^{\class}_{i}\}$  are characterised by a class $\class$. A class is one element present in BDTopo, TRI, CLC, EMA, or an Urban Unit. We define five question types that can be divided into 21 sub-questions: 
\begin{enumerate}
    \item \textbf{Presence Questions}: questions about the existence of certain objects or features in the image. 
    \begin{enumerate}
        \item \textbf{Presence}: answered with \textit{yes} if the cardinality of the set of object from the class $\numobj > 0$ and \textit{no} otherwise, where $\class \in$ BDTopo, \\e.g. \underline{"Is there a road in the image?"}
        \item \textbf{Mountain presence}  answered with \textit{yes} if the cardinality of the set of object from EMA $\numobj > 0$ and \textit{no} otherwise, \\e.g. \underline{"Are there any mountains in the image?"}
        \item \textbf{Flood presence}  answered with \textit{yes} if the cardinality of the set of objects from TRI $\numobj > 0$ and \textit{no} otherwise, e.g. \underline{"Is this area prone to flooding?"}
    \end{enumerate}

    \item \textbf{Quantity Questions}: questions about the number or amount of certain objects or features in the image.
    \begin{enumerate}
        \item \textbf{Count}: answered by $\numobj$, where $\class \in$ BDTopo, \\e.g. \underline{ "How many buildings are in the image?"}
        \item \textbf{Density} answered by $\frac{(\sum_{i=1}^{\numobj} a(\objects^{\class}_{i})) \times 100}{a(\pvhr)}$ where $a(.)$ is a function returning the area, and $\class \in$ BDTopo , \\ e.g. \underline{ "What is the religious buildings density?"}
        \item \textbf{Area}: answered by $a(\objects^{\class}_{i})$,  where $\class \in$ BDTopo ,\\ e.g. \underline{ "What is the area of the lake?"}
        \item \textbf{Percentage}: answered by $\frac{(\sum_{i=1}^{\numobj} a(\objects^{\class}_{i})) \times 100}{a(\pvhr)} $ where $\class \in$ CLC,\\ e.g. \underline{ "What percentage of the area is wetland?"}
    \end{enumerate}

    \item \textbf{Location Questions}: questions about the location of certain objects or features in the image.
    \begin{enumerate}
        \item \textbf{Absolute location}: answered by $\textrm{loc}(\objects^{\class}_{i})$ where $\textrm{loc}(.)$ is a function returning the position (both as the exact coordinate in the image space, and as the position in a $3\times 3$ grid dividing the image) of a specific object, and $\class \in$ BDTopo, \\e.g. \underline{ "Where is the largest vegetation area?"}
    \end{enumerate}

    \item \textbf{Classification Questions}: questions about the classification or type of certain objects or features in the image.
    \begin{enumerate}
        \item \textbf{Water bodies}: answered by the class $\class$ of $\max a(\objects^\class)$ or $\min a(\objects^\class)$,  where $\class \in$ BDTopo's water category (see appendix), \\e.g. \underline{ "What type of water body occupies the largest}\\ \underline{area in the image?"}
        \item \textbf{Vegetation zones} answered by the class $\class$ of $\max a(\objects^\class)$ or $\min a(\objects^\class)$,  where $\class \in$ BDTopo's vegetation category (see appendix),\\ e.g. \underline{ "Which vegetation type occupies the smallest}\\ \underline{area in the image?"}
        \item \textbf{Mountain range name}: asked only if the answer of mountain presence is \textit{yes}, and answered by $n(\objects_{i})$ where $n(.)$ is a function returning the name of the mountain range present in the geographical extent of $\pvhr$,\\ e.g. \underline{ "What is the name of the mountain range in the}\\ \underline{image?"}
        \item \textbf{Flood level}: answered by $\textrm{flood}(\epvhr)$ where $\textrm{flood}(.)$ is a function returning the highest flood risk present in $\pvhr$. The possible answers of this function are ‘low’, ‘medium’, and ‘high’. \\e.g. \underline{ "What is the flood risk level?"}
        \item \textbf{Flood type}: answered by $\textrm{type}(\epvhr)$ where $\textrm{type}(.)$ is a function returning all types of flood risks present in $\pvhr$. The possible answers of this function are 'River overflows', 'Runoff', 'Sea Flooding' and 'GroundWater overflows' \\e.g. \underline{ "What is the nature of the flood risk in this}\\ \underline{ area?"}
        \item \textbf{Land cover}: answered by the CLC class with the largest (or smallest) area,\\ e.g. \underline{ "Which land cover category occupies the}\\ \underline{largest area in the image?"}
        \item \textbf{Urban }: answered by $\textrm{urban}(\epvhr)$ where $\textrm{urban}(.)$ is a function returning the urban classification (City Center, Suburb, Isolated City, Outside Urban Unit) based on BU20,\\ e.g. \underline{ "What is the urban classification of the area}\\ \underline{in the image?"}
        \item \textbf{Department}: answered by $\textrm{dpt}(\epvhr)$ where $\textrm{dpt}(.)$ is a function returning the name of the departement of $\epvhr$, \\e.g. \underline{ "To which department does the area in the} \\\underline{image belong to?"}
        \item \textbf{Region}: answered by $\textrm{reg}(\epvhr)$ where $\textrm{reg}(.)$ is a function returning the name of the region of $\epvhr$,\\ e.g. \underline{ "To which region does the area in the image}\\\underline{belong to?"}
    \end{enumerate}

    \item \textbf{Relational Analysis Questions}:
    questions seeking to understand the relationships, comparisons, and distances between various objects or features in the image (note that for theses questions $\class \in$ BDTopo).
    \begin{enumerate}
        \item \textbf{Distance}: answered by $d(\objects^{\class_1}_{i}, \objects^{\class_2}_{j})$ where $d(.,.)$ is a function returning the distance (in meters) between two objects (note that $\objects^{\class_1}_{i}$ can be replaced by a position $\textrm{pos}$),\\ e.g. \underline{ "What is the distance between the museum}\\ \underline{and the religious place?"}
         \item \textbf{Comparison}: answered with \textit{yes} if $|\mathbf{\objects_{\class_1}}| > |\mathbf{\objects_{\class_2}}|$, \textit{no} otherwise, \\e.g. \underline{ "Are there more buildings than roads?"}
         \item \textbf{Relative location}: answered by $l^{\objects^{\class_1}_{i}}(\objects^{\class_2}_{j})$ where $l^{\objects^{\class_1}_{i}}(.)$ is a function returning the position (both as the exact coordinate in the image space, and as the position in a slice of a regular octagon) of an object with respect to $\objects^{\class_1}_{i}$,\\ e.g. \underline{ "What is the relative position of the}\\ \underline{monument with respect to the river?"}
        \item \textbf{Nearest}: answered by $l_{\textrm{pos}} (\mathbf{\objects_{\class}})$ where $l_{\textrm{pos}}(.)$ is a function returning the position (both as the exact coordinate in the image space, and as the position in a $3\times 3$ grid dividing the image) of the nearest element of a collection from the position $\textrm{pos}$,\\ e.g. \underline{ "Where is the closest road to (142, 221)?"}
    \end{enumerate}
\end{enumerate}

\begin{table*}[!t]
    \centering
    \footnotesize
    \begin{tabular}{cccccccc}
        \hline
        \textbf{Name} & \textbf{Modalities} & \textbf{Resolution} & \textbf{Annotation}& \textbf{\# Images} & \textbf{\# Questions} & \textbf{\# Q. Types} & \textbf{\# Unique A.}\\
        \hline
        TAMMI & 
        \begin{tabular}{c}BDOrtho \\ S2 - MS 10 channels \\ S1 - VV/VH/Ratio channels \end{tabular} 
        & 
        \begin{tabular}{c}0.2m \\ 10-20m \\ $5\times 20$m \end{tabular} 
        & 
        \begin{tabular}{c}  BDTopo, \\ TRI, BU20, \\ CLC, EMA \end{tabular} &
         282'852 & 3'162'514 & 21 & 109'737 
        \\ \hline
        RSVQAxBEN~\cite{lobry2021rsvqa} & S2 - RGB & 10m & BigEarthNet \cite{sumbul2019bigearthnet} & 590'326 & 14'758'150 & 2 & 26’875\\ \hline
        RSIVQA~\cite{zheng2021mutual}  & Various RGB & 0.15–8m & Manual/Other & 37'264 & 111'134 & 4 & 579  \\ \hline
        RSVQA LR~\cite{lobry2020rsvqa} & S2 - RGB & 10m & OSM & 772 & 77'232 & 4 & 9 \\ \hline
        RSVQA HR~\cite{lobry2020rsvqa} & USGS Ortho & 0.15m & OSM & 10'659 & 1'066'316 & 4 & 55\\
        \hline
    \end{tabular}
    \caption{Summary of existing RSVQA datasets, showing modalities (S1: Sentinel-1, S2: Sentinel-2), spatial resolutions, annotation sources, and dataset sizes by image and question counts.} \label{tab:Overview}
\end{table*}

\paragraph{Balancing the dataset}
One of the challenges of constructing a VQA dataset stochastically is to balance both question types and the answer types to reduce language biases~\cite{goyal2017making}.
In this work, we perform the balancing jointly at the dataset and department levels.

We perform the questions/answers pairs construction at the department level. For each VHR patch $\pvhr$ 10 questions per type are randomly constructed. We iterate through the questions created for each patch, and select questions based on four goals: 
1) having an equal number of questions for each type; 
2) having a number of question proportional to the number of VHR patches per department; 3) having a balanced distribution of answers for each type of question; 
and 4) not having more than 50 question/answer pairs for each patch.

In the case of question types with a fixed number of possible answers $\numAnswers$ such as Presence (e.g. Yes/No, $\numAnswers = 2$), the number of questions per answer is defined as 
$\numQuestionsAnswers = \frac{\numPatches}{\min(\numAnswers,10)}$, where $\numPatches$ is the number of patches in the department.
For Location questions, the locations are binned in the $3 \times{} 3$ square grid, leading to $\numAnswers=9$. 
For questions with free numerical answers such as Area or Count, the number of each of the numerical value is capped to $\numQuestionsAnswers = \frac{\numPatches}{\log(x+3)}$, where $x$ is the numerical answer and $x+3$ is chosen empirically. 
This allows us to control the distribution of each possible answer. 

Note that there is an exception for Department (3.b) and Region (3.c) questions. For each department, the answer will always be the same. To overcome this, each possible answer is capped to 2'473 for departments (the number of occurrences of the answer Paris) and 16'274 for regions (the number of occurrences of Île-de-France). This ensures a balanced distribution of the answers on the dataset level.
An overview on TAMMI and other RSVQA datasets is provided in Table \ref{tab:Overview}. It shows that the proposed dataset significantly improves on the number of question types and in diversity of answers compared to existing RSVQA datasets.

\section{Method}
\label{sec:method}
\begin{figure}[!t]
    \centering
    \includegraphics[width=\columnwidth]{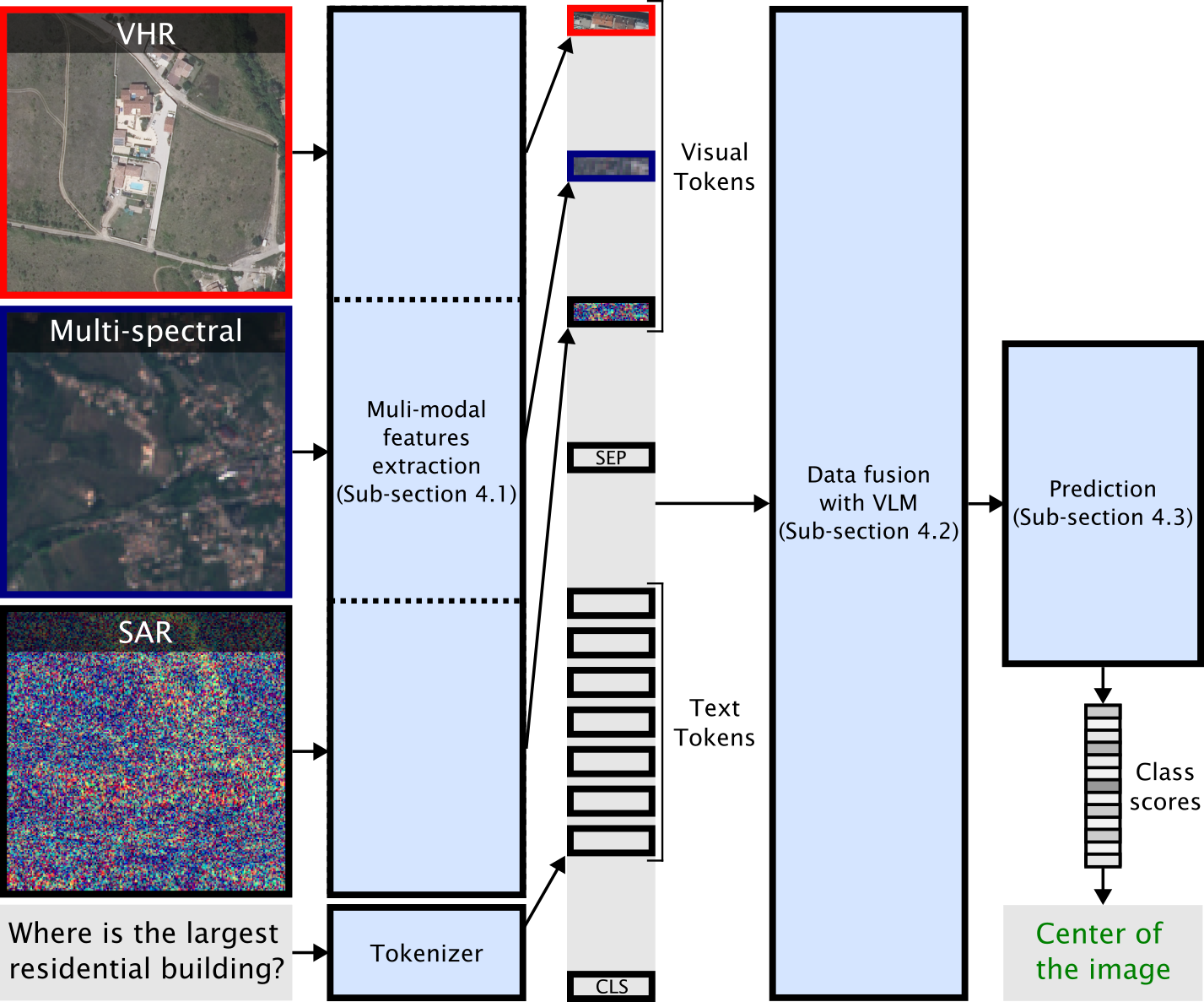}
    \caption{Graphical outline of the proposed MM-RSVQA (\textit{Multi-modal Multi-resolution} RSVQA) architecture.
    The inputs of the model (multi-modal imagery and textual question) are represented on the left and the output (predicted answer) is on the bottom right. First, we extract features from each image modality and perform a text embedding. These features are passed through a vision-language model (VLM) to obtain a vector which can be classified among a set of pre-defined answers.
    The different blocks composing the system are detailed in \autoref{sec:method}.}
    \label{fig:method}
\end{figure}

We propose a new methodology to tackle the VQA task in a multi-modal image setting. 
The general outline of the proposed architecture, named MM-RSVQA (\textit{Multi-modal Multi-resolution} RSVQA), is presented in \autoref{fig:method}.  
We first describe the feature extraction process from the different modalities (multi-modal images and questions) in \autoref{ssec:feats}. 
We then present in \autoref{ssec:fusion} the data fusion part of our model. 
Finally, the prediction of the answers is described in \autoref{ssec:pred}.

\subsection{Multi-modal features extraction}
\label{ssec:feats}
As discussed in \autoref{sec:dataset}, the questions are based on the geographical extent of the VHR patch $\pvhr$ of size $1000\times 1000$ pixels. 
Our objective is to extract relevant visual features from the $\pvhr$ patch, the MS patch $\pms$ of spatial size $\lms \times \lms$ and the SAR one $\psar$ of spatial size $\lsar \times \lsar$ to obtain useful characteristics for the VQA task.
The size of the patches $\pms$ and $\psar$ is a hyper-parameter that allows taking more or less context from the MS and SAR data into account. 
In this work, we consider the 10 bands with 10m and 20m resolution from the MS patches, and the SAR patch is the VV, VH, and the ratio VV/VH channels~\cite{tosato2024can}. Each of the $\pvhr$, $\pms$ and $\psar$ patches is passed through a separate feature extractor.


The fourth modality is the textual data corresponding to the question. The question is tokenized using DistilBERT~\cite{sanh2019distilbert}, and the tokens are converted into embeddings.

\subsection{Transformer-based data fusion}
\label{ssec:fusion}
We first project the visual features obtained through the feature extractor to a 768-dimensional vector through a linear layer. The 768-dimensional visual feature is concatenated with a [SEP] token, which marks the separation between visual and textual tokens, and then with the textual embeddings. Finally, a [cls] token is added to  represent the entire input.

To process the diverse features, we leverage VisualBERT~\cite{li2019visualbert}, a state-of-the-art vision-language model designed for the integration of visual and textual features. 
VisualBERT uses a stack of transformer layers to align regions in the images with the text input through self-attention mechanisms. 
Pre-trained on image-caption pairs, VisualBERT uses objectives as image-text matching and masked language modeling. It is effective for various tasks, such as VQA, visual commonsense reasoning, natural language for visual reasoning, and region-to-phrase grounding.

The cornerstone of the proposed method is therefore to learn jointly and in an end-to-end manner an optimized representation of the data and a fusion process. 
The latter makes it possible to exploit the complementarity of multi-modal information and to weight the best modalities, according to the visual content of the scene studied, the types of questions, and the expected answers.

\subsection{Prediction}
\label{ssec:pred}
The data fusion part of our architecture outputs a 768-dimensional vector that represents the fused multi-modal features. In this model, we frame the VQA as a classification task. Therefore, this vector is passed through a linear layer that maps it to a $k$-dimensional output.

\begin{table*}[!t]
\centering
\footnotesize
\begin{tabular}{llcccccc}
\cline{4-8}
& & & \multicolumn{4}{c}{\textbf{Ablation studies}} \\
\hline
\textbf{Category} & \textbf{Question Type} & \textbf{MM-RSVQA} & \textbf{VHR} & \textbf{MS RGB + VHR} & \textbf{MS + VHR} & \textbf{SAR + VHR} & \textbf{MS + SAR}\\
\hline
\multirow{3}{*}{\textbf{Presence Questions}} 
    & Presence & \ 96.87 & 95.41 & 96.58 & 96.79 & \textbf{97.00} & 96.84 \\
    & Flood Presence & 98.04 & 94.02 & 97.78 & 97.84 & 96.85 & \textbf{98.11} \\
    & Mountain Presence & \textbf{98.80} & 95.76 & 98.75 & 98.49 & 97.75 & 98.59 \\
\hline
\multirow{4}{*}{\textbf{Quantity Questions}} 
    & Count & \textbf{51.76} & 49.73 & 50.99 & 50.89 & 50.59 & 51.22\\
    & Density & \textbf{26.46} & 26.17 & 26.40 & 26.45 & 26.42 & 26.44 \\
    & Area & 10.28 & 10.28 & 10.30 & 10.12 & \textbf{10.33} & 10.32  \\
    & Percentage & \textbf{41.66} & 41.58 & \textbf{41.66} & 41.60 & \textbf{41.66} & 41.62 \\
\hline
\textbf{Location Questions} 
& Absolute Location & 17.17 & 17.15 & \textbf{17.21} & 16.89 & 17.04 & 17.18\\
\hline
\multirow{9}{*}{\textbf{Classification Questions}} 
    & Water & \textbf{89.38} & 81.10 & 87.73 & 88.65 & 83.14 & 88.23 \\
    & Vegetation & \textbf{53.06} & 48.57 & 50.92 & 51.00 & 48.57 & 50.83 \\
    & Flood Level & \textbf{80.92} & 75.68 & 79.19 & 79.27 & 75.68 & 79.54 \\
    & Flood Type & \textbf{97.91} & 90.60 & 95.68 & 96.93 & 95.54 & 97.32 \\
    & Land Cover & 74.76 & 68.79 & 72.53 & 72.42 & \textbf{74.89} & 74.27 \\
    & Urban & \textbf{93.78} & 69.20 & 90.33 & 91.58 & 92.40 & 93.32 \\
    & Department & \textbf{98.96} & 89.78 & 97.39 & 97.58 & 98.03 & 98.47 \\
    & Region & 99.98 & 99.96 & \textbf{99.99} & 99.92 & \textbf{99.99} & \textbf{99.99} \\
    & Mountain Name & 99.89 & 99.94 & \textbf{99.96} & 99.91 & 99.89 & \textbf{99.96} \\
\hline
\multirow{4}{*}{\textbf{Relation Questions}} 
    & Distance & \textbf{9.08} & \textbf{9.08} & 9.07 & 9.07 & 9.05 & \textbf{9.08} \\
    & Comparison & \textbf{98.43} & 98.25 & 98.34 & 98.36 & \textbf{98.43} & 98.41 \\
    & Relative Location & 17.99 & 18.10 & 17.77 & 17.62 & \textbf{18.17} & 17.74 \\
    & Nearest & 21.57 & \textbf{21.71} & 20.56 & 20.89 & 20.96 & 20.74\\
    
\hline
& \textbf{Average Accuracy} & \textbf{65.56} & 61.94 & 64.72 & 64.87 & 64.40 & 65.15 \\
& \textbf{Overall Accuracy} & \textbf{55.11} & 52.22 & 54.41 & 54.49 & 54.29 & 54.70 \\
\cline{2-8}
\end{tabular}

\caption{Accuracy for the VQA task on the TAMMI dataset.
Comparison of MM-RSVQA, VHR, MS RGB + VHR, MS + VHR, SAR + VHR, and MS + SAR across question types. Highest scores (per question type) are shown in bold.}

\label{tab:results}

\end{table*}

\section{Results and discussion}
\label{sec:results}

We present preliminary performances of the proposed method MM-RSVQA on the TAMMI dataset. With respect to the RSVQA task, we discuss the contribution (and complementarity) of the different imaging modalities with regard to the different types of questions (and answers) constituting the dataset.

\subsection{Experimental settings}
\paragraph{Dataset splitting}
The dataset is randomly split into training, validation, and test sets based on the images with a proportion of 60\%, 20\%, and 20\% respectively. We use the validation set for the tuning of the hyper-parameters described in the rest of this section. For the training, we consider only the questions having an answer in the top $k=1'000$ most frequent answers.This parameter is fixed for dimensionality reduction, as done in ~\cite{antol2015vqa}, ~\cite{lobry2020rsvqa}, covering 86.6\% of train answers. While we consider all the samples of the test set.
\paragraph{Full pipeline}
We use a frozen ResNet-152 model pre-trained on ImageNet for the orthophotos feature extractor and two ResNet-50 models pre-trained on BigEarthNet~\cite{sumbul2021bigearthnet} for the MS and SAR feature extractors. 
We set $\lms = $ 100 and $\lsar = $ 200. 
We use a cross-entropy loss, optimized with Adam. For the training of MM-RSVQA, we set the learning rate to $3\times 10^{-5}$, the batch size to 80 samples and the number of epochs to five.

\subsection{Metrics}
Three metrics are used to evaluate the VQA results: the per-class accuracy, overall accuracy (OA) and average accuracy (AA). 
The per-class accuracy is defined as the ratio of correct answer with the total number of questions for one of the 21 question types. 
The overall accuracy is the ratio of correct answers with the total number of questions in the dataset. 
Finally, the average accuracy is the arithmetic mean of the per-class accuracies.

\subsection{Quantitative results}
To evaluate the proposed architecture and the TAMMI dataset, we conduct experiments using our main model MM-RSVQA and we perform ablation studies with various combinations of modalities. 
The results are presented in Table \ref{tab:results}. We show the performances of the MM-RSVQA model (using VHR, MS and SAR modalities) and ablation studies across question types, highlighting the contributions of each modality in the context of this challenging task.

\paragraph{MM-RSVQA} 
The proposed multi-modal multi-resolution baseline model, as seen in Table \ref{tab:results}, achieves strong performance across most question types and outperforms models using only one or two modalities. This highlights the advantage of integrating multiple modalities for the RSVQA task. 
Notably, MM-RSVQA demonstrates high accuracy through categories such as presence, comparison, water and region. 
These results indicate that combining VHR, MS, and SAR data helps the model to better identify objects, to assess quantities, and to improve classification accuracy. 
For some question types, such as absolute location, area and relative location, the model shows lower accuracy. This suggests that these questions types are more challenging due to the difficulty of understanding spatial relationships or precisely locating objects. One hypothesis is that the added trainable parameters when adding MS and SAR modalities requires more training samples than for other models. 

\paragraph{Ablation study} 
To evaluate the contribution of the different modalities we perform ablations studies at the input level. The results, shown in Table \ref{tab:results} present the accuracy of different configurations: VHR only; MS RGB + VHR (only keeping the RGB bands of Sentinel-2 and VHR patch); MS + VHR (10 bands of Sentinel-2 and VHR patch); SAR + VHR; and MS + SAR (Sentinel-1 SAR images + 10 bands of Sentinel-2).

From these results we observe that using VHR only does not allow to obtain good results. Despite the fact that the questions only concern the geographical extent of the VHR patch, it appears that providing additional context, either through MS or SAR, brings better performances. This is clearly visible in classification questions such as department or urban. This suggests that the integration of SAR and multi-spectral data enhances the ability of the model to understand and classify complex features. This validate the main hypothesis of this work: additional context, even if the question is restricted to a set geographical extent, improve performances.

By comparing MS RGB + VHR and MS + VHR, we can see that the performances are similar, with a small improvement for the model considering the 10 spectral bands of Sentinel-2 despite the added parameters. This validate the approach taken in other datasets considering Sentinel-2 data (RSVQA LR, RSVQAxBEN, see \autoref{tab:Overview}) which only considered the RGB channels. However, to the best of our knowledge, it is the first time that this hypothesis is experimentally demonstrated for VQA.

Regarding the SAR modality, we can see that SAR + VHR obtains similar performances to MS + VHR. This indicates that the benefits of adding context holds whether the modality. However, it appears that SAR is particularly efficient at discriminating certain features, such as land cover. Finally, our experiment with MS + SAR modalities indicates that jointly considering both modalities is a strong advantage for the model. Indeed, despite not having the very high resolution data, this model obtains the second best overall performances behind MM-RSVQA.

\section{Conclusion and limitations}
In this work, we address the VQA on Remote Sensing data problem by utilizing multi-modal and multi-resolution data. With these diverse modalities, we show that a model can benefit from their complementarity, in terms of coverage, spectral resolution, spatial resolution and physical information. 
This represents a new challenge for the vision community, as the interaction between different image modalities, beyond RGB, and text remains under-explored.  To support this new problem, we propose a new dataset, named TAMMI, which contains 21 types of questions based on very high resolution patches (orthophotos), high-resolution multispectral data (Sentinel-2), and Synthetic Aperture Radar (Sentinel-1) images. The dataset spans three French departments, each with distinct landscapes, allowing for diverse and challenging questions. 

We introduce a baseline model to process multi-modal images and textual input data based on the VisualBERT architecture. Our results indicate that multi-resolution and multi-modal data enhance model performance. The VHR modality is useful for tasks requiring high-detail images. For tasks requiring a broader context, the contributions from the SAR and MS modalities prove to be substantial. In our pipeline, we use all three modalities together. However, ablation studies show that even when one modality is missing, the proposed model obtains better performances compared to VHR RGB orthophotos alone. This is a strong result, as this modality is the one commonly used in RSVQA.


Future work includes developing better models using specialized feature extractors for multispectral and SAR data. This dataset introduces a challenging task and, thanks to dynamic context selection, supports a variety of multi-modal applications beyond VQA. It is also easily extensible to cover more regions, promoting better generalization across diverse landscapes and environments.

{
    \small
    \bibliographystyle{ieeenat_fullname}
    \bibliography{main}
}
\clearpage
\setcounter{page}{1}
\maketitlesupplementary

\section*{SAR data (Sentinel-1) pre-processing}
\begin{figure}[!ht]
    \centering
    \begin{subfigure}[b]{0.49\columnwidth}
        \centering
        \includegraphics[width=\textwidth,trim={1.5cm 1.5cm 1.5cm 1.5cm},clip]{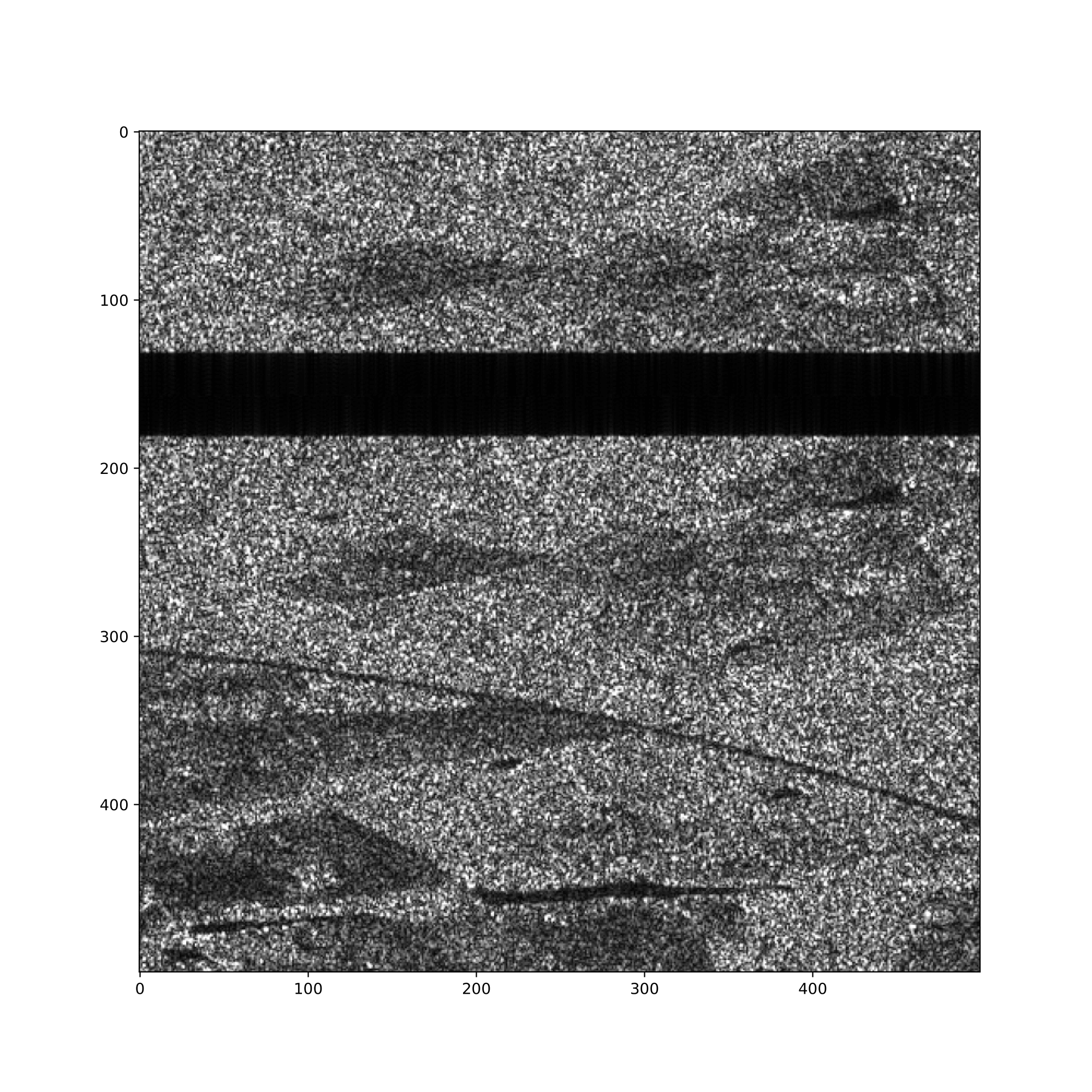}
        \caption{Pre-debursting}
        \label{fig:sub1}
    \end{subfigure}
    \begin{subfigure}[b]{0.49\columnwidth}
        \centering
        \includegraphics[width=\textwidth,trim={1.5cm 1.5cm 1.5cm 1.5cm},clip]{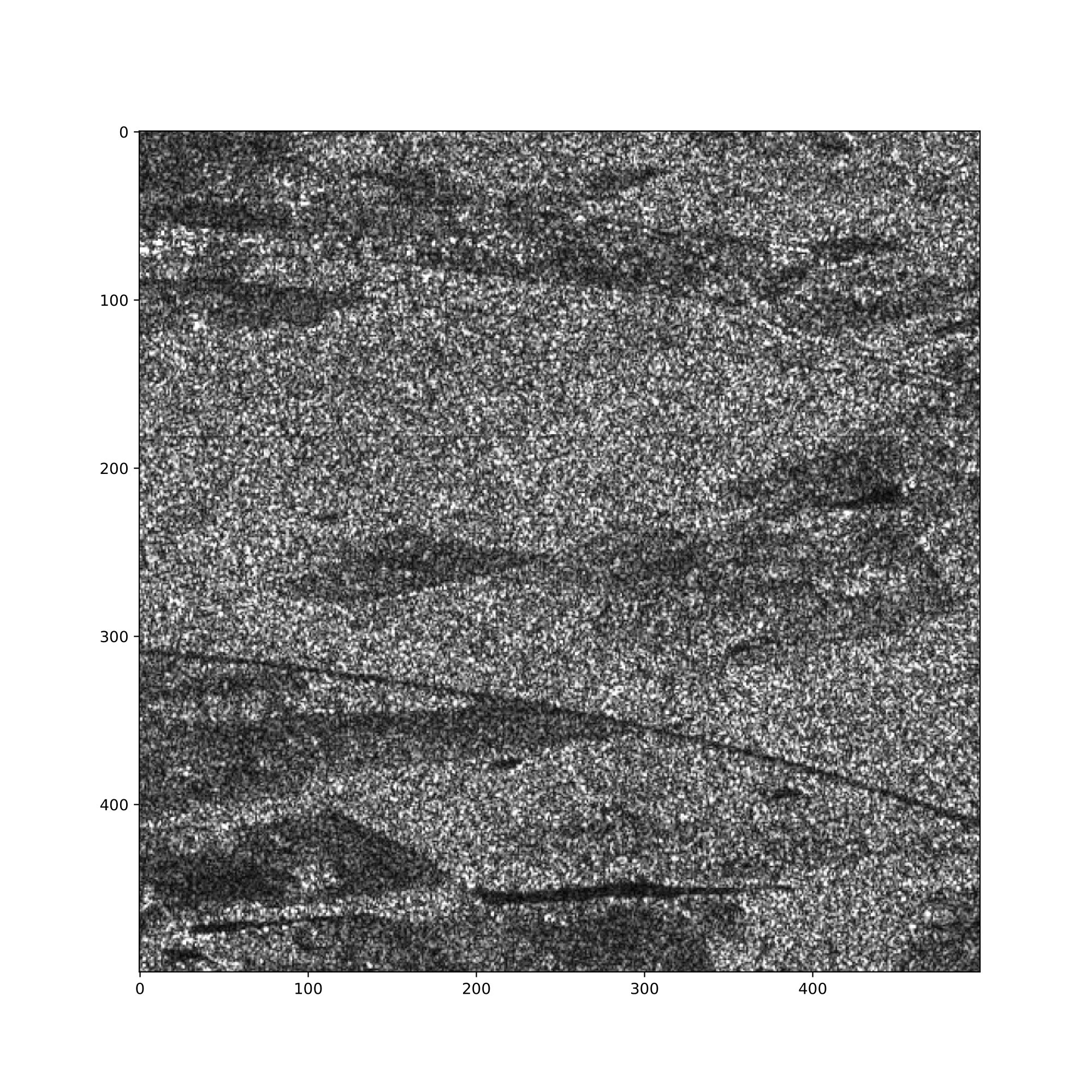}
        \caption{Post-debursting}
        \label{fig:sub2}
    \end{subfigure}
    \caption{Sentinel-1 SLC images before (a) and after (b) the application of the proposed debursting method. 
    The visualization is done using a threshold of 233.}
    \label{fig:sarprepostdeburst}
\end{figure}

The pre-processing pipeline to generate continuous SAR (Sentinel-1) patches consists of several steps including SAR and VHR data. 
We describe hereinafter each step in details:
\begin{enumerate}
    \item We use the algorithm presented in~\cite{weissgerber2022labsar} to find the position of the center of the VHR patch in the SAR image. To do so, we need the geographical position of the center of the VHR patch. This position must include altitude, due the geometrical distortions inherent in SAR, particularly layover;
    
    \item The latitude and longitude of the VHR patches are extracted from the meta-data of the image. Using this latitude and longitude, the altitude is given by this \href{https://geoservices.ign.fr/documentation/services/services-deprecies/calcul-altimetrique-rest}{https://geoservices.ign.fr/documentation/services/services-deprecies/calcul-altimetrique-rest} and these information are stored in a JSON file, \textit{latlonalt.json};
    
    \item The Sentinel-1 Single Look Complex (SLC) images are separated in three swath. To find the correct swath, the projection of the geographical point is applied, using the meta-data linked to each swath. The only swath for which the algorithm returns a valid position is selected;
    
    \item In each swath, the S1 image is divided in different overlapping bursts, that are all stored in the same file separated by black lines. 
    The S1 images need to be debursted (removing of the black line and of the overlap) to get a continuous image before to extract the S1 patch that is inputted in the model. 
    This debursting is done using meta-data of the S1 image and by comparing the two consecutive burst. A result of the debursting process is shown in Figure \ref{fig:sarprepostdeburst}.
    The debursted images are provided with the dataset. The projection algorithm is thus modified to give the position in this new deburst file. This position and the correct swath are stored in the JSON file containing all information on the VHR patches;
    
    \item When a patch of size $\lsar$ $\times$ $\lsar$ ($\lsar$ specified by the user) is extracted, the two polarimetric channels are converted in dB, and the ratio is computed. A tail-value elimination procedure is performed on each channel separately using statistics information extracted over the whole dataset.
\end{enumerate}

\newpage
\clearpage

\end{document}